\def\year{2021}\relax
\documentclass[letterpaper]{article} 
\usepackage{aaai21}  
\usepackage{times}  
\usepackage{helvet} 
\usepackage{courier}  
\usepackage[hyphens]{url}  
\usepackage{graphicx} 
\usepackage{natbib}  
\usepackage{caption} 
\usepackage{subcaption}
\usepackage{multirow}
\usepackage{lipsum}
\usepackage{multicol}
\usepackage{amsmath}
\usepackage{comment}
\usepackage{tabularx}
\usepackage{booktabs}
\usepackage{paralist}

\newcommand\gridworld{\textsc{GridWorld}}

\newcommand\cpmdp{\textsc{Cp-Mdp}}
\newcommand\candecompparafac{\textsc{Candecomp-Parafac}}
\newcommand\tabularpi{\textsc{Tabular-PI}}
\newcommand\tabularvi{\textsc{Tabular-VI}}
\newcommand\cpmdppi{\textsc{Cp-Mdp-PI}}
\newcommand\cpmdpvi{\textsc{Cp-Mdp-VI}}
\usepackage[colorinlistoftodos, 
	]{todonotes}

\newtheorem{definition}{Definition}

\title{\cpmdp{}: A CANDECOMP-PARAFAC Decomposition Approach \\to Solve a Markov Decision Process Multidimensional Problem}

\author{
    Daniela Kuinchtner,\textsuperscript{\rm 1}
    Afonso Sales,\textsuperscript{\rm 2}
    Felipe Meneguzzi\textsuperscript{\rm 3} 
    \\
}
\affiliations{
    Pontifical Catholic University of Rio Grande do Sul (PUCRS)\\
    daniela.kuinchtner@edu.pucrs.br,\textsuperscript{\rm 1}
    afonso.sales@pucrs.br,\textsuperscript{\rm 2} 
    felipe.meneguzzi@pucrs.br\textsuperscript{\rm 3}

}

\begin{document}

\maketitle

\begin{abstract}
Markov Decision Processes (MDPs) are the underlying model for optimal planning for decision theoretic agents in stochastic environments.
Although much research focuses on solving MDP problems both in tabular form or using factored representations, none focused on tensor decomposition methods. 
Solving MDPs using tensor algebra offers the prospect of leveraging advances in tensor-based computations to further increase solver efficiency. 
In this paper, we develop an MDP solver for a multidimensional problem using a tensor decomposition method to compress the transition models and optimize the value iteration and policy iteration algorithms. 
We empirically evaluate our approach against tabular methods and show our approach can compute much larger problems using substantially less memory, opening up new possibilities for tensor-based approaches in stochastic planning. 
\end{abstract}

\section{Introduction}

\noindent Markov Decision Processes (MDPs) have been widely studied as an elegant mathematical formalism to model stochastic domains~\cite{bellman_MDP_1957}. 
The solution to an MDP problem, given a stochastic state transition system and a reward function, is a policy that defines the optimal (or maximum expected utility) action for every state in the domain. 
Most approaches to solve tabular representations of these problems require a large number of mathematical operations and substantial memory. 
Such tabular approaches, while mathematically sound, have limited applicability, because of the \emph{curse of dimensionality}, when the required computational resources scale exponentially with the number of state variables~\cite{book-dynamic-programming-bellman}. 

Subsequent research developed methods to factorize the transition model, such as Factored MDPs~\cite{boutilier_policy_1995,Guestrin_efficient_algorithms_factored_mdp_2003,karinaDelgado_solver_factered_MDP_2011}, and produce compact representations of complex uncertain systems allowing exponential reduction in representation complexity. 
Such factored approaches are grounded on representing the transition model as dynamic Bayesian networks (DBNs). 
Methods to solve DBN-based representations, however, do not leverage advances in tensor decomposition methods to represent large MDPs or to improve solver runtime. 
In this paper, we develop an efficient tensor representation for $n$-dimensional problems characterised by a small number of parameters to represent the transition models, enabling solvers to scale up and mitigate the curse of dimensionality. 
We call the resulting approach \emph{\cpmdp{}}, addressing the challenges of (1) reducing the necessary memory and the computational cost required by tabular methods to compute the solution, and (2) leveraging advances in tensor processing to further increase solver efficiency of MDP solvers by representing the transition models as tensor components.
Our main contributions are: (1) an expanded formalization of~\cite{tensor-based-mdp-daniela} 
to multidimensional model-based MDP problems formalization; and (2) a novel implementation of the \gridworld{} using the \candecompparafac{}~\cite{CANDECOMP&INDSCAL} decomposition method.

\section{Background}

\subsection{Markov Decision Process}

A Markov decision process is a sequential decision problem for a fully observable and stochastic environment with a Markovian transition model. 
An MDP consists as a tuple $\mathcal{M} = \langle \mathcal{S}, \mathcal{A}, \mathcal{P}, \mathcal{R}, \gamma \rangle$~\cite[Ch. 3]{sutton_book_reinforcement_learning_2018}, where:
\begin{inparaitem}
    \item[] $\mathcal{S}$ is the state space;
    \item[] $\mathcal{A}$ is the action space;
    \item[] $\mathcal{P}$ is a transition probability function {$\mathcal{P}(s' \mid s, a)$};
    \item[] $\mathcal{R}$ is a reward function; and
    \item[] $\gamma \in [0..1]$ is a discount factor.
\end{inparaitem}
Most MDP solvers seek to find a policy that assigns an (optimal) action choice for each agent at each state~\cite{Guestrin_PhDThesis_2003}. 
A policy with highest expected reward or with expected reward equal to all other policies is called the optimal policy ({$\pi^{*}$}).

\emph{Value Iteration} and \emph{Policy Iteration} are common dynamic programming algorithms to solve MDPs. The basic idea of value iteration is to calculate the utility of each state and then use the state utilities to select an optimal action in each state \cite{article-dynamic-programming-bellman}.
The utility of a state is the immediate reward for this state plus the expected discounted utility of neighboring states, assuming the agent chooses the optimal action. 
The \textit{Bellman Equation}~\cite{bellman_MDP_1957} formalizes this model in Equation~\ref{eq:utility}. Value iteration propagates information through the state space iteratively by means of local updates until it converges to the optimal value, from which the optimal policy is extracted. 

\begin{equation}\label{eq:utility}
    \displaystyle V(s) = \mathcal{R}(s) + \gamma  \max_{a \in \mathcal{A}(s)} \sum_{s'} \mathcal{P}(s' \mid s, a)V(s')
\end{equation}
Unlike value iteration, the execution of policy iteration alternates in two steps: policy evaluation and policy improvement. 
Policy evaluation is the process of computing the value function for a fixed policy $\pi$. 
Whereas policy improvement is the process of generating a new policy, such that $V_{\pi'}(s) \geq V_{\pi}(s)$, by acting greedily with respect to $\pi$~\cite[Ch. 4]{sutton_book_reinforcement_learning_2018}. 
The algorithm terminates when the policy improvement step yields no change in the policy.
At this point, the policy is a solution to the Bellman equation, and $\pi$ must be an optimal policy.

\subsection{Tensor Decomposition}

A tensor is a multidimensional array, where the $n^{th}$-order tensor is an element of the tensor product of $n$ vector spaces, each of which has its own coordinate system~\cite{kolda_tensor_decomposition_applications_2009}. 
We apply tensor decomposition to data arrays to extract and explain their properties, and we can consider these methods to be higher-order generalizations of matrix singular value decomposition (SVD) and principal component analysis (PCA)~\cite{kolda_tensor_decomposition_applications_2009}.
Increasing computing capacity has enabled tensor-based approaches to decompose higher-order problems, leading to a number of applications in signal processing, computer vision, data mining, neuroscience and machine learning~\cite{kolda_tensor_decomposition_applications_2009,sidiropoulos_tensor_decomposition_signal_ML_2017}.

\begin{figure}[h]
  \centering
  \includegraphics[width=0.45\textwidth]{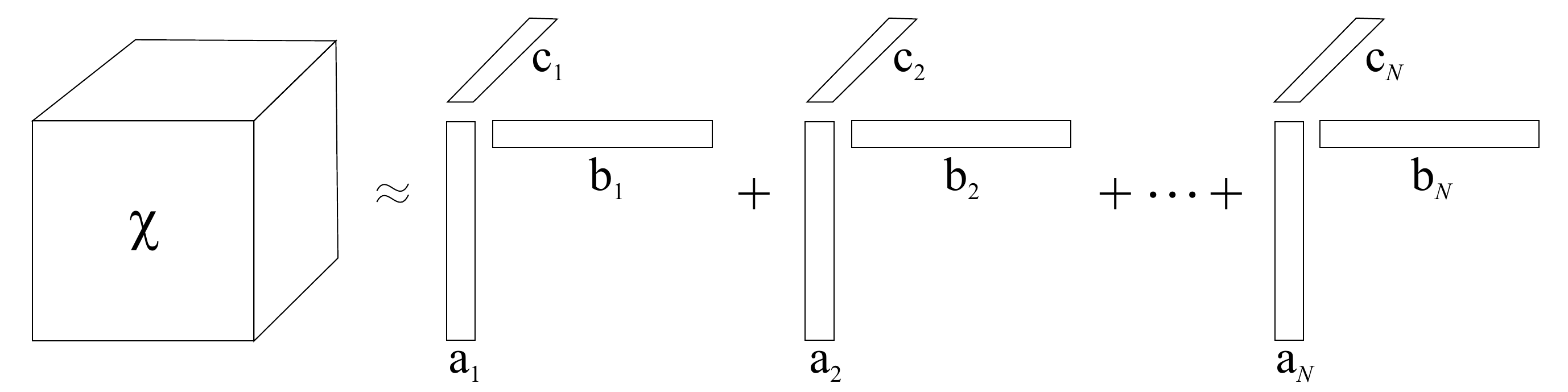}
  \caption{\label{fig:tensor-cp}A \candecompparafac{} decomposition.}
\end{figure}

Canonical Polyadic Decomposition with Parallel Factors, also known as \candecompparafac{} decomposition or CP decomposition~\cite{CANDECOMP&INDSCAL,PARAFAC2,Kiers-CP} is an example of tensor decomposition method. 
\candecompparafac{} is the process that factorizes a tensor into sums of individual components, providing a parallel proportional analysis and an idea of multiple axes for analysis~\cite{kolda_tensor_decomposition_applications_2009}. This method expresses a tensor as sum of the outer product of vectors, and we illustrate a \candecompparafac{} decomposition of a third-order tensor $\mathcal{X}$ in Figure~\ref{fig:tensor-cp}.

\section{Tensor Decomposition of MDPs}
In order to improve solution's runtime and memory usage, our method decomposes transition models into small tensor components, by using the \candecompparafac{} decomposition key idea as the basic semantics of this work: to express the tensor as a sum of rank-one tensors. 

\subsection{MDP Tensor-based Formalization}

We now formalize MDPs in terms of $n$-dimensional values, which can be applied to any discrete, stochastic and finite MDP problem. 

\begin{definition}\label{def:dimensions}
Let $\mathcal{D}$ be the set of environment's dimensions, where $\mathcal{D}$ = \{$d_1$, $d_2$, $d_3$, ..., $d_D$\} and $D = |\mathcal{D}|$\footnote{The notation adopted is $|\mathcal{X}|$ to define the cardinality of a set.} is the number of dimensions. 
Each component of this set can be analyzed individually and comprises a set of states as follows:
\begin{itemize}
    \item[$d_1$] set of states of dimension 1, where $d_1$ = \{$s_{1_{d_1}}$, $s_{2_{d_1}}$, $s_{3_{d_1}}$, ..., $s_{S_{d_1}}$\} and $S_{d_1} = |d_1|$ is the number of states of dimension 1;\\
    \item[$d_2$] set of states of dimension 2, where $d_2$ = \{$s_{1_{d_2}}$, $s_{2_{d_2}}$, $s_{3_{d_2}}$, ..., $s_{S_{d_2}}$\} and $S_{d_2} = |d_2|$ is the number of states of dimension 2;\\
    ...
    \item[$d_D$] set of states of dimension $D$, where $d_D$ = \{$s_{1_{d_D}}$, $s_{2_{d_D}}$, $s_{3_{d_D}}$, ..., $s_{S_{d_D}}$\} and $S_{d_D} = |d_D|$ is the number of states of the dimension $D$.
\end{itemize}
\end{definition}

\begin{definition}\label{def:statespace}
Given the set of dimensions $\mathcal{D}$, it contains $|\mathcal{D}|$ dimensions $d$, respectively, where $d$ $\in [1..D]$, and the statespace $\mathcal{S}$ is set of all states of all dimensions (such that $S = |\mathcal{S}|$), where:
\begin{align*}
\mathcal{S} = 
\{ &
(s_{1_{d_1}}, s_{1_{d_2}}, ..., s_{1_{d_D}})_{s_{1}}, ..., (s_{1_{d_1}}, s_{1_{d_2}}, ..., s_{2_{d_D}})_{s_{2}},\\
& (s_{1_{d_1}}, s_{1_{d_2}}, ..., s_{3_{d_D}})_{s_{3}}, ...,(s_{S_{d_1}}, s_{S_{d_2}}, ..., s_{S_{d_D}})_{s_{S}} \}
\end{align*}
\end{definition}

Finally, we define the \candecompparafac{} tensor components representation by $\mathcal{C}_{(s)}$:

\begin{definition}\label{def:tensor-components}
$\mathcal{C}_{(s)}$ is the set of tensor components of each state $s$, where $C_{(s)}$ = $|\mathcal{C}_{(s)}|$ is the number of tensor components of each state $s$:
\begin{align*}
\mathcal{C}_{(s)} =  
\{ & 
[(s, s', \mathcal{P}(s'\mid s,a)), ... , (s, s', \mathcal{P}(s'\mid s,a))]_{C_{s1}},  ..., \\
& [(s, s', \mathcal{P}(s'\mid s,a)), ... , (s, s', \mathcal{P}(s'\mid s,a))]_{C_{s2}},  ..., \\
& [(s, s', \mathcal{P}(s' \mid s,a)), ..., (s, s', \mathcal{P}(s'\mid s,a))]_{C_{sS}} \} 
\end{align*}

We define each component by: a) a current state $s$; b) a successor state $s'$; and c) a probability of state transition $\mathcal{P}(s'\mid s,a)$.
The number of tensor components $C_s$ of each state is that of transitions with $\mathcal{P}(s'\mid~s,a) > 0$.
\end{definition}

\subsection{MDP Scenario and Problem Statement}

We evaluate our approach using standard \gridworld{} problems, where grid size is defined by the number of states $\mathcal{S}$. 
The number of actions is determined $2*\mid\mathcal{D}\mid$, i.e., we consider two actions for each Cartesian plane.
Obstacles $\mathcal{O}$ are states an agent has no access to.  
The interaction with the environment terminates when the agent reaches one of the terminal states $\mathcal{T}$.
For example, for a two-dimensional grid, the environment is defined by $x \times y$ states and the agent's actions in a given state are \textit{North} and \textit{South} in x-axis; and in y-axis are \textit{West} and \textit{East}. 
For a three-dimensional grid, the environment is defined by $x \times y \times z$ states and the actions are \textit{Forward} and \textit{Backward} in the z-axis; and so on. 

\section{Experimental Results}
\begin{table*}[htb]
\centering
\begin{tabular}{rr|rrrrr}
\toprule

&\textbf{$\mathcal{D}$}& \multicolumn{1}{r}{\textbf{2} } & \multicolumn{1}{r}{\textbf{3} }  & \multicolumn{1}{r}{\textbf{5} } &  \multicolumn{1}{r}{\textbf{7} }& \multicolumn{1}{r}{\textbf{9} }\\ 
\midrule
&\textbf{$\mathcal{A}$}& \multicolumn{1}{r}{\textbf{4} } & \multicolumn{1}{r}{\textbf{6} } & \multicolumn{1}{r}{\textbf{10} } &  \multicolumn{1}{r}{\textbf{14} }& \multicolumn{1}{r}{\textbf{18} }\\ 
\midrule
\textbf{$\mathcal{T}$} & \textbf{$\mathcal{O}$}  &&\multicolumn{3}{l}{\textbf{Number of States}}\\
\midrule
        6   &50      &4,900	    &4,000       &3,125        &2,048	    &3,888       \\
        8   &100     &10,000     &8,000   	&7,000	     &5,184	    &5,832      	\\
        10  &200     &14,400	    &12,500	    &10,000	     &9,216	   &8,748	    		\\
        12  &300     &19,600	    &18,750		&12,500	     &10,368     &9,216     \\
        14  &400     &22,500     &24,000	    &19,200      &18,432&17,496	    	    	\\
        16  &500     &90,000	    &60,000	    &100,000	    &78,125&82,944	    	    	\\
        18  &600     &250,000	&125,000     &200,000	    &233,280&196,008	    	   	\\
		20  &700     &640,000	&512,000	    &600,000	     &605,052&491,520        	\\
		22  &800     &1,000,000	&1,000,000	&1,200,000	  &823,543&1,000,000      	\\
\bottomrule
\end{tabular}
\caption{Grid configuration.}
\label{tab:configuration}
\end{table*}

We evaluate the Python implementation of our approach\footnote{Available at: https://github.com/danielakuinchtner/cp-mdp} 
using an Intel(R) Xeon(R) CPU @ 2.30GHz with 13 GB RAM from a Python 3 Google Colaboratory notebook\footnote{https://colab.research.google.com/}.
We compare our results against a standard tabular implementation of value iteration and policy iteration algorithms available in Python $pymdptoolbox$\footnote{https://pypi.org/project/pymdptoolbox/}. 
The plotted values consist of the average of 6 executions of the \gridworld{} problem for each grid size configuration.
We use a \emph{discount factor} of $\gamma = 0.9$ and $maxIter = 1,000$. We define non-terminal states with a -3 reward, and set terminal states with an additive +100 reward or a discounted -100 reward.
The grid contains randomly placed obstacles and terminal states.
Table~\ref{tab:configuration} shows the number of actions $\mathcal{A}$, states $\mathcal{S}$, obstacles $\mathcal{O}$ and terminals $\mathcal{T}$ we consider for each test environment. 
For example, for a 2D test with 4 actions, we use 6 terminals and 50 obstacles to solve a 4,900-state problem, where the total number of states are divided into two dimensions (e.g. 70$\times$70 = 4,900).

\begin{figure*}[htb]
 \centering
 \begin{subfigure}[]{0.48\textwidth}
 \centering
 \includegraphics[width=\textwidth]{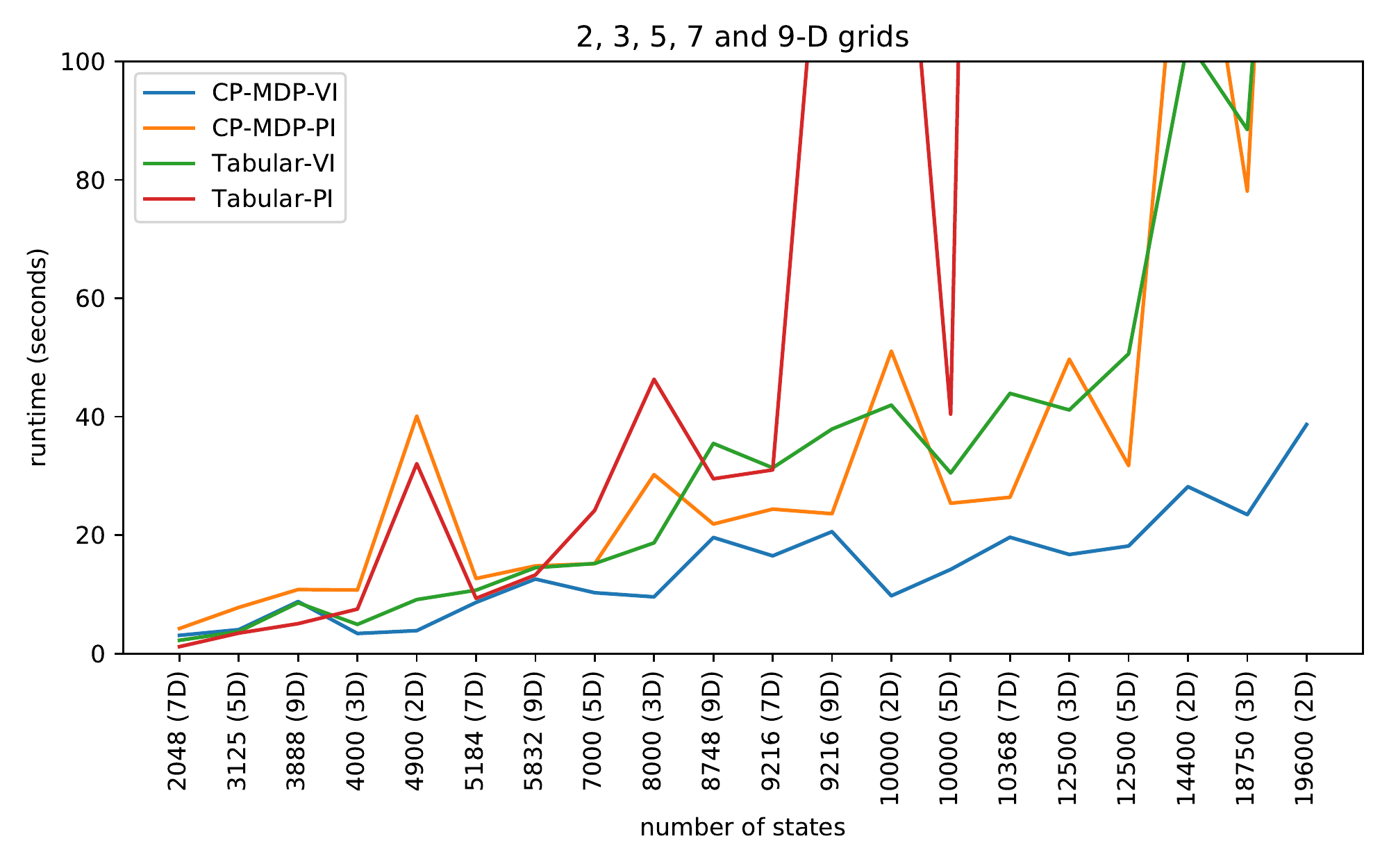}
 \caption{Runtime (in seconds)}
 \label{fig:runtime}
 \end{subfigure}
 \begin{subfigure}[]{0.48\textwidth}
 \centering
 \includegraphics[width=\textwidth]{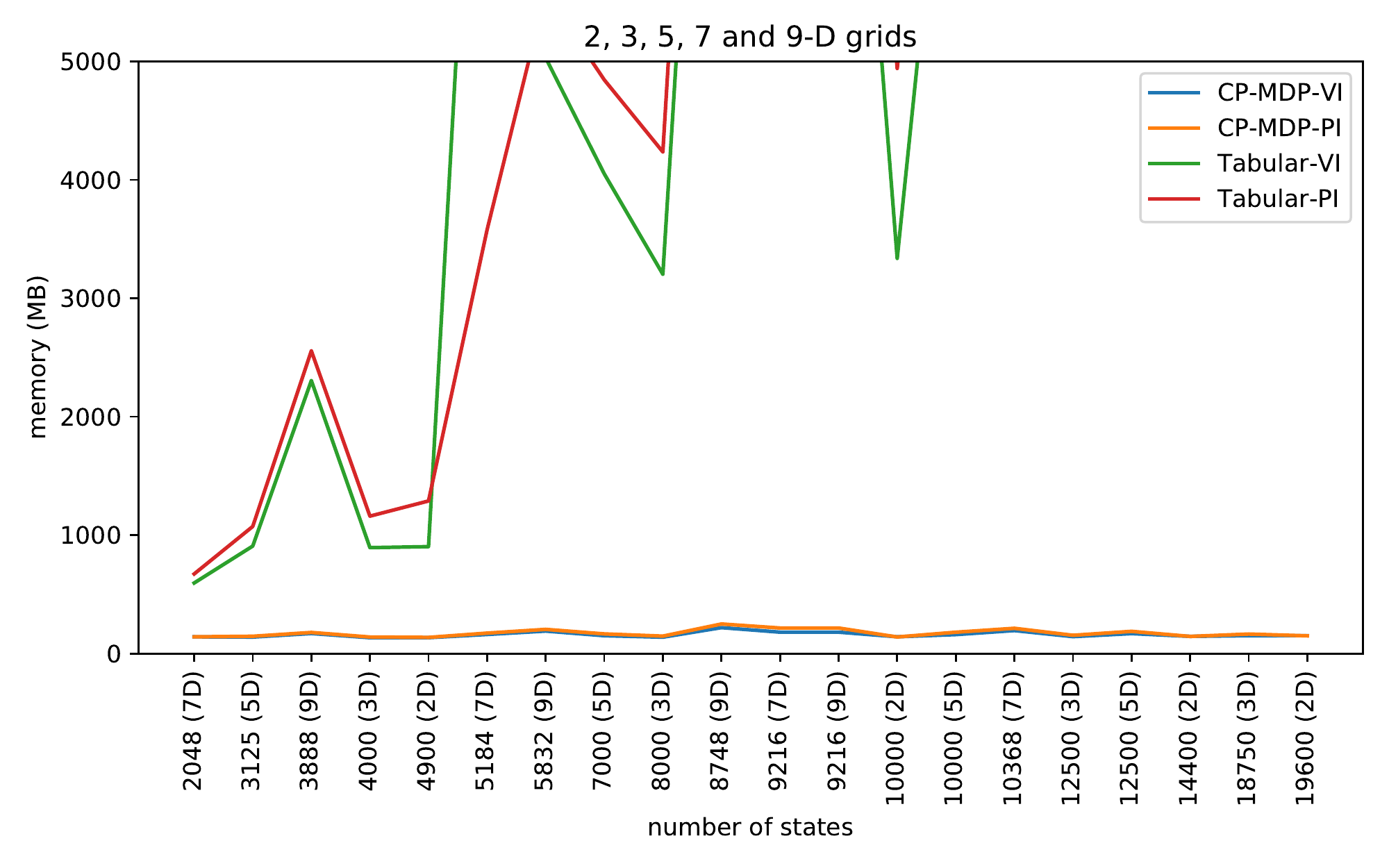}
 \caption{Memory (in MB)}
 \label{fig:memory}
 \end{subfigure}
 \caption{(a) Runtime (in seconds) and (b) memory (in MB) of \cpmdpvi{} and \cpmdppi{} methods against the tabular value iteration (\tabularvi{}) and policy iteration (\tabularpi{}) algorithms of 2, 3, 5, 7 and 9 dimensions.}
 \label{fig:memory-runtime}
\end{figure*}

Figure~\ref{fig:runtime} shows the runtime (in seconds) to solve the \gridworld{} problem using the value iteration (\tabularvi{}) and policy iteration (\tabularpi{}) algorithms both with a tabular representation and our \cpmdp{} value iteration (\cpmdpvi{}) and \cpmdp{} policy iteration (\cpmdppi{}) methods for problems with 2, 3, 5, 7 and 9 dimensions. 
While \cpmdp{} does not significantly outperform tabular methods for smaller problems (up to 10,000 states), it achieves a runtime improvement between 60\% and 80\% in most larger cases. 

Nevertheless, as we show in Figure~\ref{fig:memory}, the memory requirements for the \cpmdp{} method is substantially lower than tabular ones, to the extent that we cannot run problems with more 20,000 states. 
For a 19,600-state test, \tabularvi{} uses 12.45 GB and  \tabularpi{} uses more memory than our set limit of 13 GB, whereas both \cpmdpvi{} and \cpmdppi{} compute the solution using 151.33 MB and 149.66 MB, respectively. 
Consequently, the memory improvement of \cpmdp{} is more than 90\% for all cases with more than 5,000 states compared to tabular value and policy iteration. 

In most cases, \tabularpi{} is substantially less efficient than \tabularvi{}, and \tabularpi{} cannot solve problems with much fewer states and a larger number of dimensions (e.g. $> 10,000$ states and 9 dimensions).
This behavior is due to the computational cost of both methods, since the complexity of \tabularvi{} is $O(|S|^2 \times |A|)$, whereas the complexity for \tabularpi{} is $O(|S|^3 + |S|^2 \times |A|$, where $S$ is the number of states and $A$ is the number of actions. 

\begin{figure*}[htb]
 \centering
 \begin{subfigure}[]{0.48\textwidth}
 \centering
 \includegraphics[width=\textwidth]{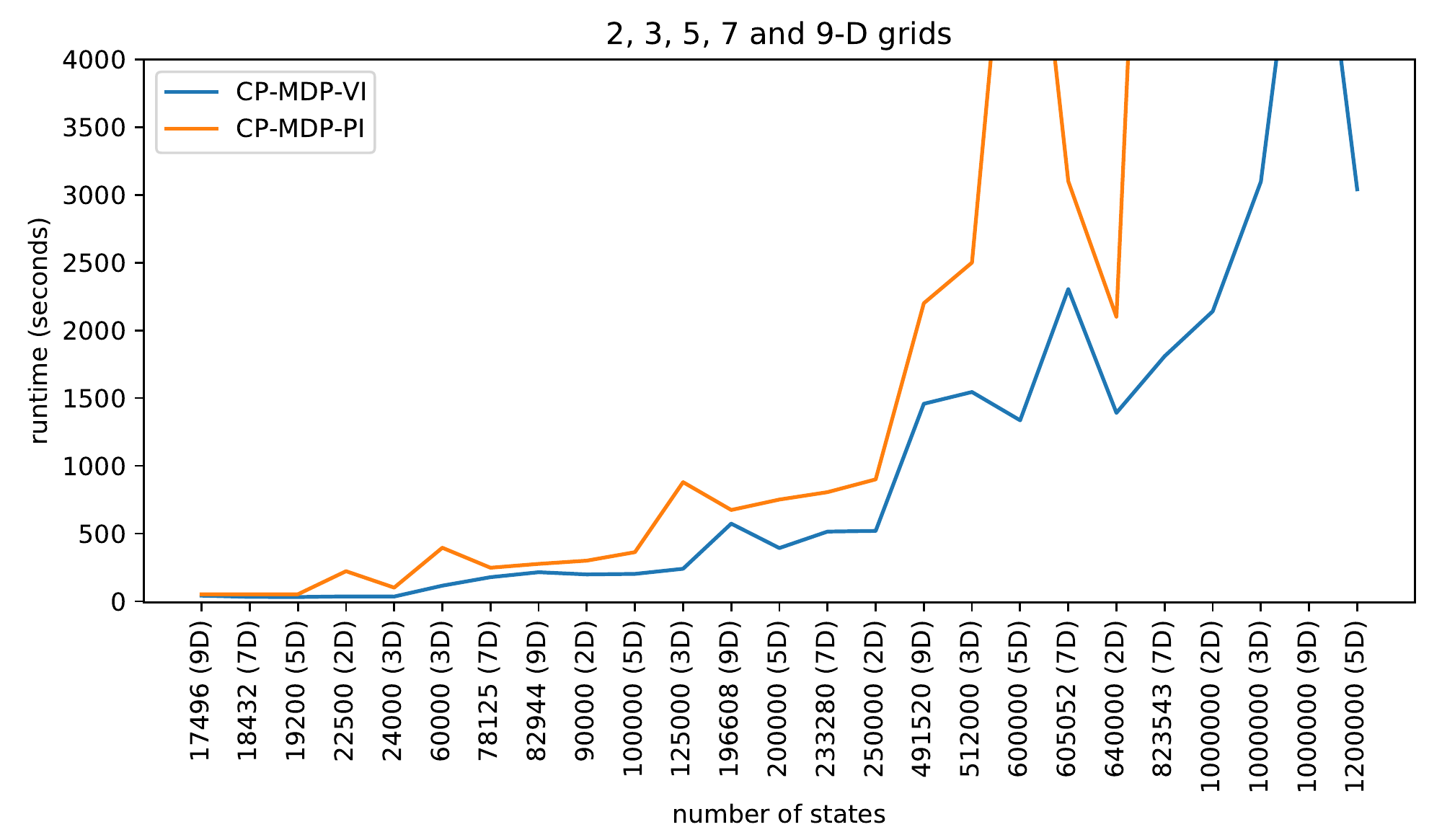}
 \caption{Runtime (in seconds)}
 \label{fig:runtime-cp}
 \end{subfigure}
 \begin{subfigure}[]{0.48\textwidth}
 \centering
 \includegraphics[width=\textwidth]{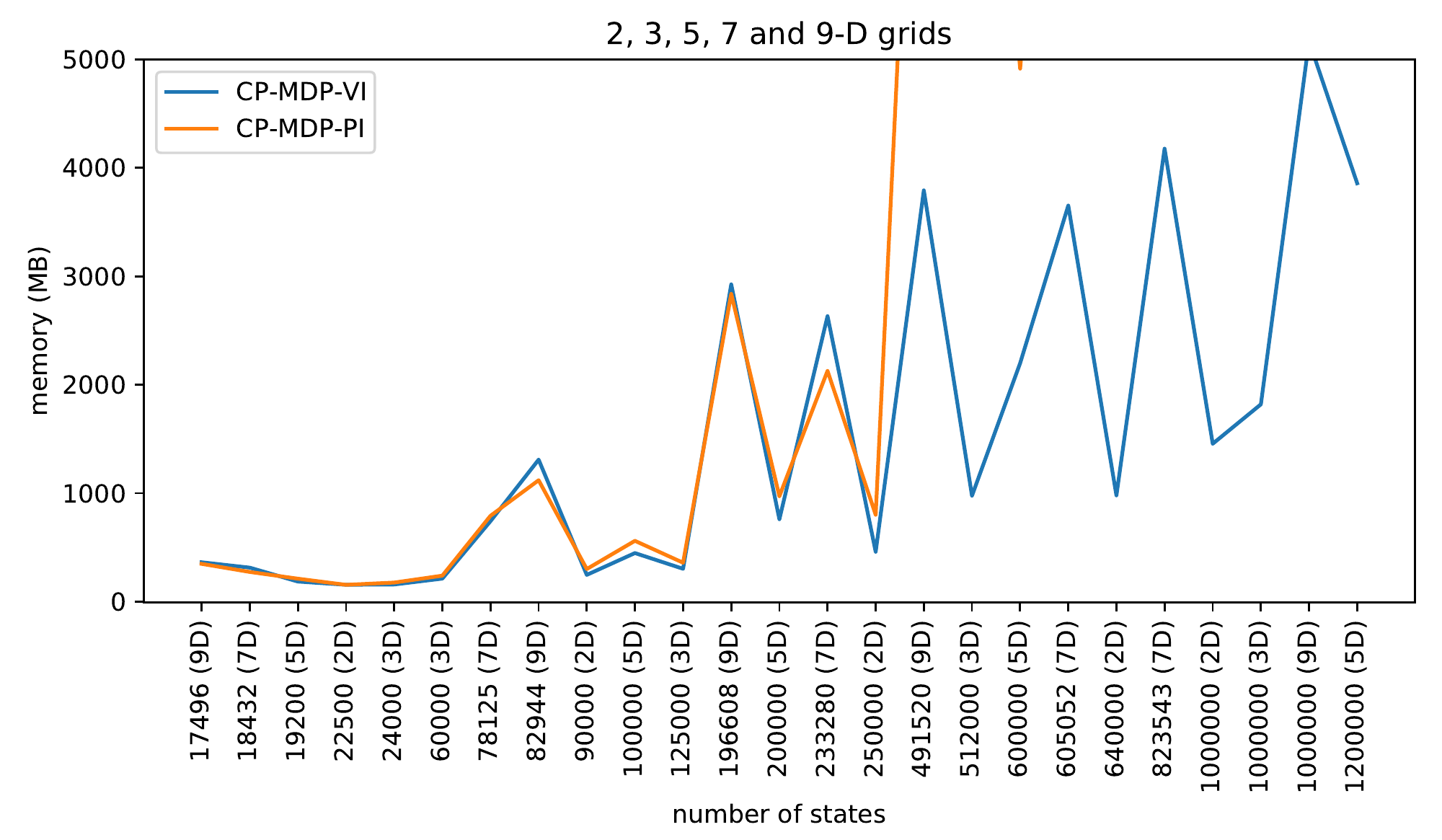}
 \caption{Memory (in MB)}
 \label{fig:memory-cp}
 \end{subfigure}
 \caption{\cpmdp{} runtime (a) and necessary memory (b) to compute large grid sizes using the value and policy iteration.}
 \label{fig:memory-runtime-cp}
\end{figure*}

Figures~\ref{fig:runtime-cp} and~\ref{fig:memory-cp} show runtime and memory results for tests performed only by the \cpmdp{} method, due to memory limitations to compute the solution for these grid sizes using tabular approaches. 
In terms of memory, \cpmdpvi{} and \cpmdppi{} are very similar for smaller grid sizes, but for more than 500,000 states, \cpmdpvi{} requires 50\% or less memory than \cpmdppi{}.
For problems with 2 or 3 dimensions, \cpmdpvi{} runtime and memory improve by approximately 65\% compared to higher-dimensional problems with the same number of states.  
For example, \cpmdpvi{} computes the solution of a 1,000,000-state 9-dimensional problem in 5,895.82 seconds using 5.16 GB, however, \cpmdpvi{} takes 2,140.34 seconds using 1.45 GB to solve a problem with the same number of states over 2 dimensions.

Finally, in order to compare our \cpmdpvi{} and \cpmdppi{} against tabular methods, we compute their computational cost. 
Since \cpmdpvi{} requires a number of multiplications equal to:
$\prod_{i=1}^{S} s_i \times \prod_{j=1}^{C_{(s)}} c_j \times \prod_{k=1}^{A} a_k$, 
\noindent we arrive at a complexity of $O(|S| \times |C_{(s)}| \times |A|)$, where $C_{(s)}$ is the number of tensor components for the state $s$.
By contrast, \cpmdppi{} requires a number of multiplications equal to:
$\prod_{i=1}^{S} s_i \times  \prod_{j=1}^{C_{(s)}} c_j  \times  \prod_{k=1}^{C_{(s)}} c_k + \prod_{l=1}^{S} s_l \times \prod_{m=1}^{C_{(s)}} c_m \times \prod_{n=1}^{A} a_n$,
\noindent resulting in a $O(|S| \times |C_{(s)}|^2 + |S| \times |C_{(s)}| \times |A|)$ complexity. 
In consequence, these methods require much less computation, and substantially less memory than traditional tabular methods.

\section{Conclusion and Further Work}

We developed a tensor decomposition method, called \cpmdp{}, which uses the idea of the \candecompparafac{} decomposition, to solve an MDP multidimensional problem using value iteration and policy iteration algorithms. To our knowledge our work is the first one employing tensor decomposition to solve MDPs.
Our empirical analysis shows that \cpmdpvi{} method solve more efficiently than \cpmdppi{} and tabular methods in both runtime and memory.
First \cpmdpvi{} achieves runtime improvements of up to 80\% in the best cases compared to tabular approaches. 
Second, both methods require substantially less memory to compute these solutions, decreasing memory use by more than 90\% for large multidimensional problems. 

In order to leverage advances in GPU computation libraries, we tried several different ways to parallelize our approach to run on GPUs with Tensorflow\footnote{https://www.tensorflow.org/}, Pytorch\footnote{https://pytorch.org/} and Numba\footnote{https://numba.pydata.org/}, however the runtime did not improve in any of these libraries. 
Such negative result seems to stem from the communication overhead between CPU and GPU, as our methods rely on multiplications between small tensor components. 

As future work, we intend to use other methods of tensor decomposition, such as Tensor-Train Decomposition~\cite{TT-decomposition} and Tucker Decomposition~\cite{tucker} to perform the transition models factorization, aiming to improve runtime on GPUs. 
We also intend to generalize the \cpmdp{} implementation to solve several types of problems already described in Relational Dynamic Influence Diagram Language (RDDL)~\cite{rddl-scott-sanner}, aiming to compare our approach against the state-of-art of factored MDPs solvers for different problems.

\clearpage
\bibliography{bib}

\end{document}